\documentclass[runningheads]{llncs}
\usepackage{multirow}
\usepackage{wrapfig}
\usepackage[T1]{fontenc}
\usepackage{amsmath}
\usepackage{booktabs}
\usepackage {amssymb}
\usepackage{colortbl}
\usepackage{xcolor}
\usepackage{multirow}
\usepackage{graphicx}
\usepackage{bbding}
\usepackage{hyperref}
\usepackage{orcidlink}
\usepackage{graphicx}
\begin{document}
\title{FDDet: Frequency-Decoupling for Boundary Refinement in
Temporal Action Detection}
\titlerunning{FDDet}
%
\author{Xinnan Zhu\orcidlink{0009-0009-6955-6238} \and
Yicheng Zhu\orcidlink{0009-0000-5432-6784} \and
Tixin Chen\orcidlink{0009-0004-5179-4530} \and
Wentao Wu\orcidlink{0009-0003-7352-7997} \and
Yuanjie Dang\Envelope\orcidlink{0000-0002-8302-1338}} 
\authorrunning{Z. Zhu et al.}
%
\institute{College of Computer Science and Technology, Zhejiang University of Technology, Hangzhou 310023, China \\
\email{\{zxn, zeeyc, chentx, 202203150422, dangyj\}@zjut.edu.cn}}
\maketitle              
\begin{abstract}
Temporal action detection aims to locate and classify actions in untrimmed videos. While recent works focus on designing powerful feature processors for pre-trained representations, they often overlook the inherent noise and redundancy within these features. Large-scale pre-trained video encoders tend to introduce background clutter and irrelevant semantics, leading to context confusion and imprecise boundaries. To address this, we propose a frequency-aware decoupling network that improves action discriminability by filtering out noisy semantics captured by pre-trained models. Specifically, we introduce an adaptive temporal decoupling scheme that suppresses irrelevant information while preserving fine-grained atomic action details, yielding more task-specific representations. In addition, we enhance inter-frame modeling by capturing temporal variations to better distinguish actions from background redundancy. Furthermore, we present a long-short-term category-aware relation network that jointly models local transitions and long-range dependencies, improving localization precision. The refined atomic features and frequency-guided dynamics are fed into a standard detection head to produce accurate action predictions. Extensive experiments on THUMOS14, HACS, and ActivityNet-1.3 show that our method, powered by InternVideo2-6B features, achieves state-of-the-art performance on temporal action detection benchmarks.
\end{abstract}
\section{Introduction}
Temporal Action Detection (TAD) is a crucial task in long-form video understanding, aiming to locate action instances by predicting their start time, end time, and corresponding categories in untrimmed videos. 

From the perspective of training, existing TAD methods can be broadly classified into three categories: 1) Feature-based methods, such as TriDet~\cite{tridet}, treat pre-trained video encoders as fixed and use pre-extracted features, significantly reducing computation and simplifying training. However, the lack of adaptability in frozen features may lead to suboptimal localization performance. 2) Full fine-tuning methods~\cite{etad,afsd} update the entire pre-trained model to learn task-specific representations, offering better alignment with TAD tasks but requiring substantial computational resources, which limits practicality. 3) Adapter-based methods, such as AdaTAD~\cite{adatad}, insert lightweight trainable modules while freezing the backbone, achieving a balance between efficiency and adaptability. Nonetheless, they still require loading the full backbone during training, making them slower than feature-based methods.

\begin{figure}[t] 
    \centering
    \includegraphics[clip, trim=0.1cm 0.5cm 0.1cm 0.5cm,width=0.8\columnwidth]{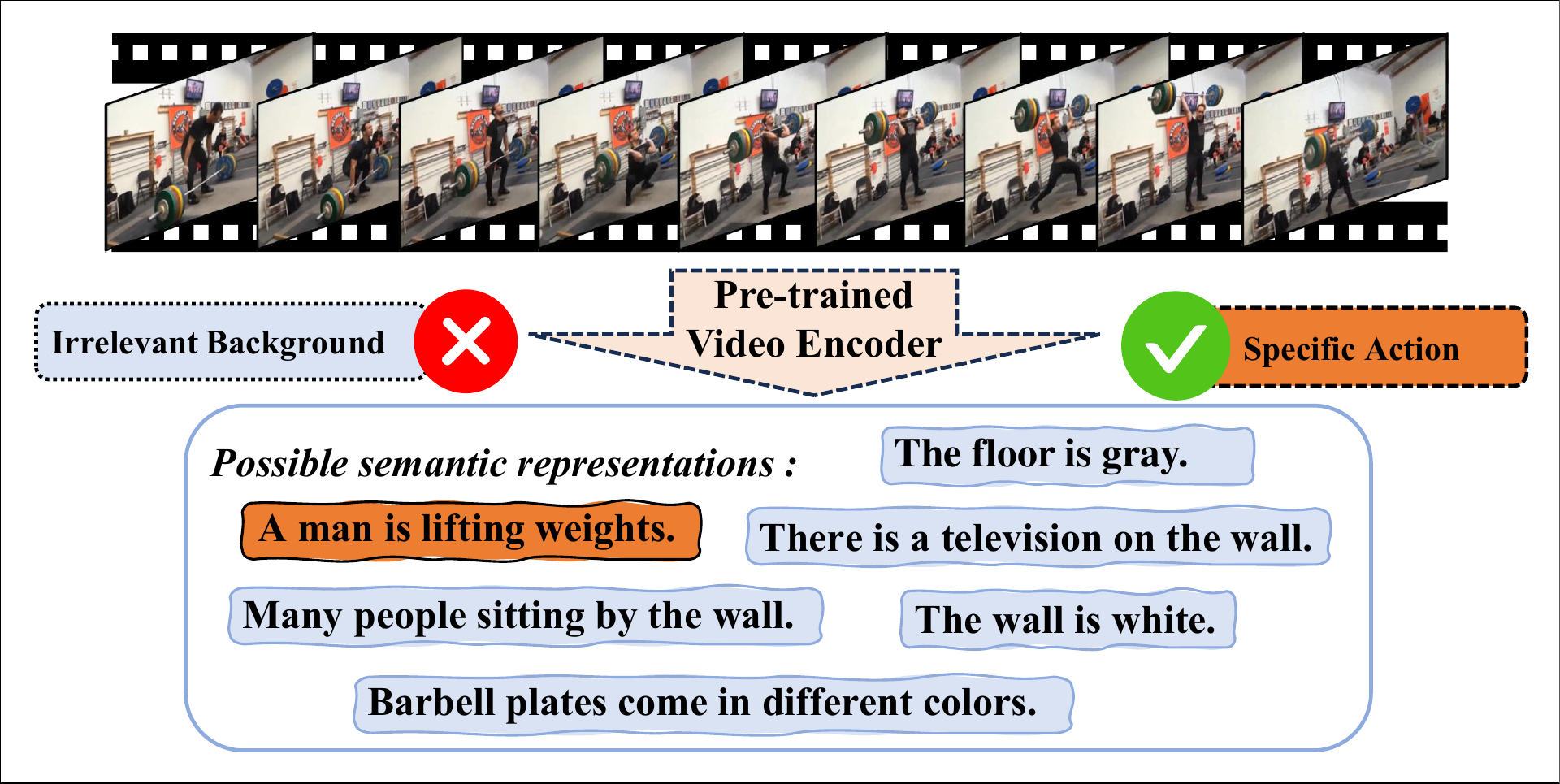} 
    \caption{This figure illustrates how pre-trained video encoders capture excessive background details, such as “The floor is gray” or “There is a television on the wall,” along with task-relevant features like “A man is lifting weights.”}
    \label{fig:actionfre}
\end{figure}

Feature-based methods remain the most prevalent approach in TAD due to their efficiency and simplicity. These methods leverage pre-trained video encoders to extract features from untrimmed videos, keeping them fixed during training and directly applying them for action detection. However, the reliance on fixed features limits the model’s ability to adapt its representations to the specific requirements of TAD tasks. Since pre-trained encoders are optimized for general video understanding rather than action detection, they often capture excessive background information, leading to context confusion and imprecise temporal boundaries. As illustrated in Figure~\ref{fig:actionfre}, these models encode not only high-level semantic representations but also numerous background details, such as subtle textures, background motions, and scene transitions. While these details may be useful for general video analysis, they can be detrimental to TAD by dominating the extracted representations and emphasizing trivial background elements. This over-representation of background context reduces the model’s ability to focus on action-relevant features, resulting in degraded action localization. Specifically, it introduces two major challenges: 1) the suppression of critical action cues due to the overwhelming presence of background signals, and 2) increased ambiguity in temporal boundary prediction, as models struggle to differentiate between actual action occurrences and irrelevant scene changes. Consequently, background interference obscures crucial action signals, leading to lower detection accuracy and a higher rate of false positives.

To this end, we propose a novel frequency-aware decoupling network that enhances the discriminability of action representations by filtering out cluttered semantic details captured by pre-trained models. Specifically, we design an adaptive temporal decoupling scheme that dynamically suppresses irrelevant semantics while preserving fine-grained atomic action details, ensuring a more refined and task-specific feature representation. Furthermore, we enhance interframe feature modeling by capturing temporal variations, allowing the model to better distinguish action dynamics from background redundancy and improve motion transition representation. To further refine temporal boundaries and mitigate localization ambiguity, we introduce a long-short-term category-aware relation network that captures both fine-grained local transitions and long-range dependencies. By integrating these components, our framework achieves superior action-background separation and more precise boundary predictions.

In summary, our contributions are as follows:
\begin{itemize}
    \item We propose a frequency-guided atomic action decoupling network that improves action representation by suppressing background interference, enhancing temporal structures, and refining semantic information. This approach facilitates better separation of action and background, ensures stable action representations over time, and enhances the adaptability of extracted features to TAD tasks. To our knowledge, this is the first study to incorporate frequency domain analysis into TAD tasks.

    \item We propose the long-short-term category-aware relation network (TCAR), which captures both fine-grained local transitions and long-range dependencies. This module significantly improves the localization of temporal boundaries while reducing ambiguity, leading to more accurate action localization.

    \item Extensive experiments on THUMOS14~\cite{thumos}, HACS~\cite{hacs}, and ActivityNet-1.3~\cite{activitynet} demonstrate that our method, leveraging InternVideo2-6B~\cite{internvideo2} features, achieves state-of-the-art performance in TAD tasks. Our results validate the effectiveness of frequency-aware action decoupling in refining action representations, improving temporal boundary precision, and enhancing overall detection accuracy.
\end{itemize}

\section{Related Work}
\subsection{Temporal Action Detection}
Temporal Action Detection aims to localize and classify actions in untrimmed videos. Existing methods can be broadly categorized into feature-based and end-to-end frameworks: the former decouples feature extraction and detection by using pre-trained extractors, while the latter\cite{etad,adatad} jointly optimize video encoders and detectors for better task-specific representation learning. Based on detection strategy, TAD methods are also classified into one-stage\cite{actionformer,actionmamba,tridet,dyfadet}, two-stage\cite{vsgn,gtad,bmn}, and DETR-based\cite{tadtr,dualdetr} approaches. Among them, one-stage methods have recently gained popularity due to their efficiency and competitive performance. Representative works include ActionFormer\cite{actionformer}, which employs transformer encoders to capture long-range dependencies; TriDet\cite{tridet}, which introduces triplet point modeling for more accurate boundary localization; and DyFADet\cite{dyfadet}, which dynamically aggregates multi-scale features to adapt to variable action durations. Despite their strengths, these methods mainly focus on architectural innovations while ignoring the redundancy inherently present in pre-extracted video features. To address this issue, we propose FDDet, a novel one-stage detector that introduces frequency domain decoupling to eliminate redundant components and enhance informative patterns, thereby improving both efficiency and detection accuracy.

\subsection{Video pre-trained Models} 
Video pre-trained models are fundamental to Temporal Action Detection (TAD), as they provide rich semantic and temporal features learned from large-scale video datasets. Popular models such as I3D\cite{i3d}, VideoMAEv2\cite{videomaev2}, and InternVideo2 have shown strong generalization to downstream tasks. Among them, InternVideo2\cite{internvideo2}, with its fine-grained and multimodal feature representations, is particularly effective at capturing detailed visual and temporal information. However, these models are typically trained for general-purpose video understanding tasks like classification or retrieval, which leads them to encode not only task-relevant action cues but also a large amount of background content, scene context, and object-level details. This results in feature redundancy, especially in temporal dimensions, where subtle but irrelevant variations (e.g., lighting changes, repetitive context) are preserved. Such redundancy can blur action-relevant signals and hinder precise temporal localization. To address this, we propose a frequency-guided action representation framework that suppresses background interference, highlights temporal structures, and refines semantic features, ultimately enhancing the adaptability of pre-trained features to the specific demands of TAD.

\subsection{Frequency Domain Analysis}
Frequency domain analysis has shown strong potential in capturing temporal patterns by transforming signals into the frequency space, and is widely used for denoising, anomaly detection, and structural enhancement in traditional signal processing. Recently, its ability to reveal periodicity and suppress noise has attracted interest in video understanding tasks\cite{fre}. However, directly applying such techniques to TAD remains challenging due to the complex and non-stationary nature of video features. In particular, the Discrete Fourier Transform (DFT) provides a global perspective on temporal frequency components, but struggles to capture fine-grained local variations critical for precise action boundary localization in untrimmed videos. To address these limitations, we propose the FGAAD network, which incorporates local frequency cues as a complement to global modeling. By jointly considering global structures and rapid local changes, FGAAD enhances the representation of action boundaries and improves the robustness of temporal localization.

\subsection{State Space Model}

State Space Models (SSMs)\cite{ssm,mamba} capture temporal dynamics by modeling sequences as state transitions, where hidden states evolve over time based on prior states and input observations. Although SSMs are effective in modeling both short- and long-term dependencies, their application to TAD poses challenges. In particular, the global focus of hidden state updates can dilute moment-specific information, making it difficult to precisely align transitions with action boundaries. This misalignment can hinder accurate localization and reduce sensitivity to fine-grained temporal changes. To address this, we enhance the SSM structure with boundary-aware modeling to strengthen temporal alignment and improve action localization performance.

\section{Method}
\begin{figure*}[ht] 
    \centering
    \includegraphics[clip, trim=0.1cm 1.0cm 0.1cm 1.0cm, width=\columnwidth]{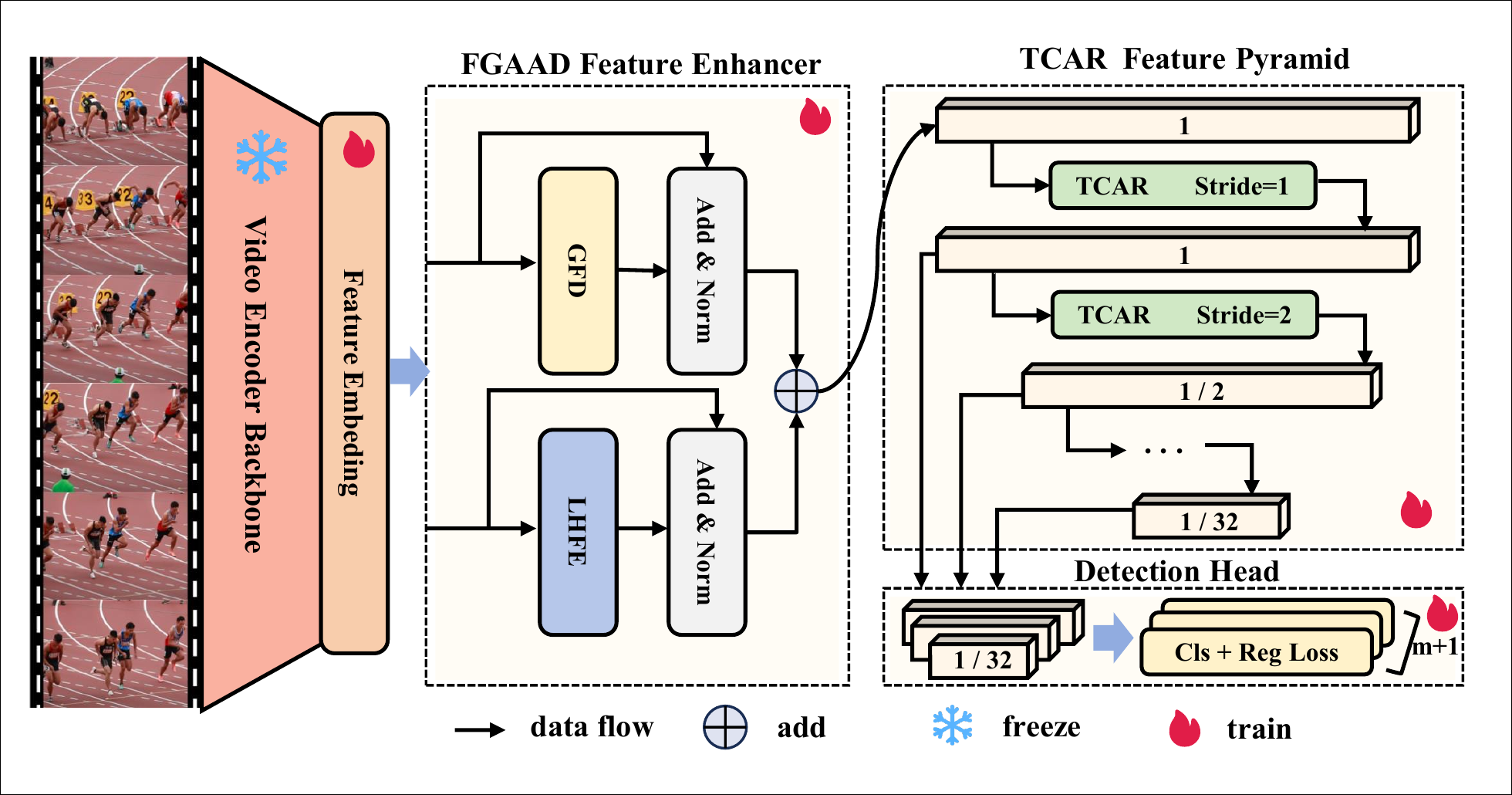}
    \caption{Overview architecture of FDDet.}
    \label{fig:overview}
\end{figure*}
\subsection{Problem Definition}
An untrimmed video \( X \) is represented as a sequence of feature vectors \( X = \{x_1, x_2, \dots, x_T\} \), where \( T \) denotes the total number of time steps. Each \( x_t \), typically extracted by pre-trained models such as I3D\cite{i3d} and SlowFast\cite{slowfast} for 3D convolutional networks, or VideoMAE\cite{videomaev2} for Transformer-based architectures, represents a video clip at time \( t \).  
The goal of TAD is to predict a set of action instances \( \Psi = \{\psi_1, \psi_2, \dots, \psi_N\} \), where \( N \) is the number of actions. Each instance \( \psi_n = (s_n, e_n, a_n) \) includes the start time \( s_n \), end time \( e_n \), and action category \( a_n \) from a pre-defined set of \( C \) categories, subject to \( s_n < e_n \).
\subsection{Method Overview}
We propose a simple yet effective one-stage temporal action detection framework, as illustrated in Fig.~\ref{fig:overview}. The framework consists of four key components: a feature extraction backbone, a feature decoupling enhancer, a temporal feature refinement module, and a boundary detection head. First, video features are extracted using a pre-trained backbone network and stored locally to avoid redundant feature extraction in later training stages. These features are then processed by the decoupling enhancer to filter out irrelevant noise. The refined features are passed through a temporal feature refinement module, which builds a multi-level feature pyramid. Finally, the pyramid is fed into the classification and regression heads for precise action localization and detection.
\subsection{Frequency-Guided Atomic Action Decoupling Network}  
Our proposed Frequency-Guided Atomic Action Decoupling (FGAAD) network is designed to enhance action localization by leveraging frequency-domain cues. It consists of two key components: Global Frequency Decoupling (GFD) and Local High-Frequency Enhancement (LHFE), which work collaboratively to filter out irrelevant patterns and emphasize fine-grained, action-relevant features. As illustrated in Fig.~\ref{fig:fgaad}, the FGAAD network operates on pre-extracted features \( x \in \mathbb{R}^{B \times L \times D} \), where \( B \) denotes the batch size, \( L \) represents the temporal length of each sequence, and \( D \) is the feature dimension obtained from pre-trained models. 

\subsubsection{Global Frequency Decoupling}

To capture temporal dynamics in the frequency domain, we apply the Discrete Fourier Transform (DFT) along the temporal dimension of the extracted features:

\begin{equation}
s_x[k] = \sum_{n=0}^{L-1} x[n] \cdot e^{-i \frac{2\pi k n}{L}}, \quad k = 0, 1, \dots, L-1,
\label{eq:dft}
\end{equation}
where \( s_x \in \mathbb{C}^{B \times L \times D} \) is the complex-valued frequency spectrum of the input sequence. Each frequency index \(k\) corresponds to a temporal oscillation with frequency \( \omega_k = \frac{2\pi k}{L} \), where lower values of \(k\) capture slow, stable temporal trends, and higher values capture rapid, localized variations.

\begin{figure*}[t] 
    \centering
    \includegraphics[clip, trim=0.1cm 0.5cm 0.1cm 0.5cm, width=1\columnwidth]{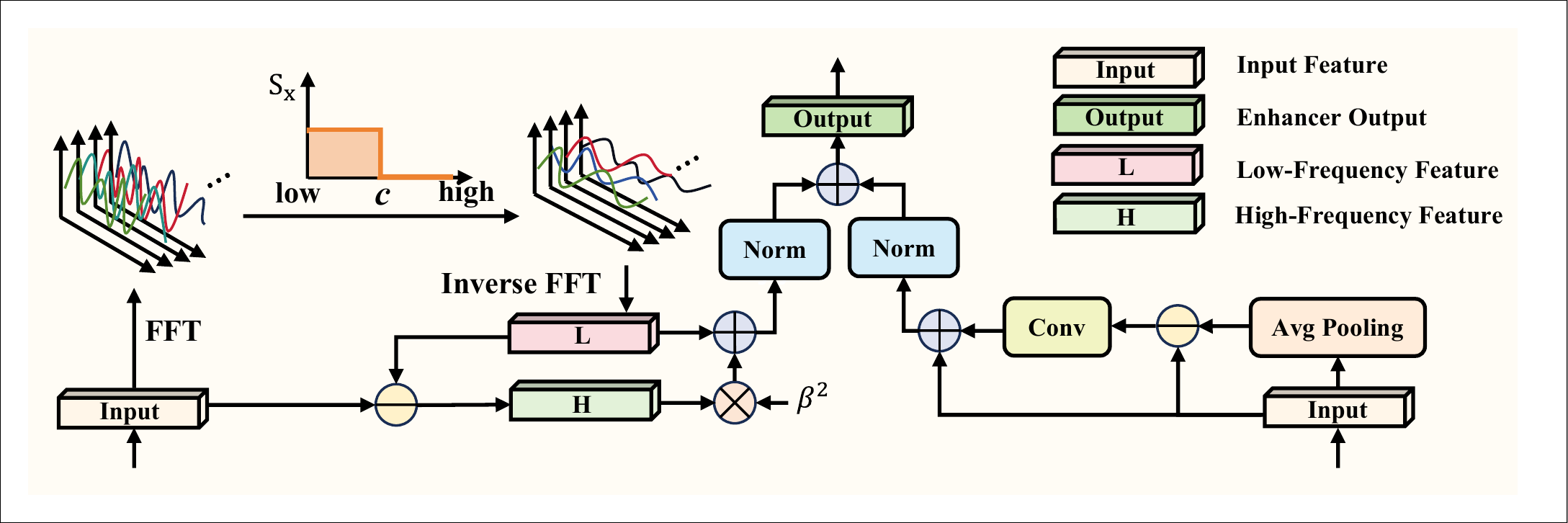}
    \caption{Overview architecture of FGAAD Block. Here \(\beta\) is a learnable parameter. We use a stage-specific cutoff frequency c in the frequency domain to separate high and low frequencies, helping the model focus on meaningful temporal patterns.}
    \label{fig:fgaad}
\end{figure*}

To extract coarse-grained temporal patterns, we apply a low-pass filter that retains only the first \(c\) frequency components, where \(c\) is a hyperparameter defining the frequency cutoff, as illustrated in Fig.~\ref{fig:fgaad}. The filtered spectrum \(s_{x,\text{L}}\) is then transformed back into the time domain using the Inverse Discrete Fourier Transform (IDFT):

\begin{equation}
\text{L}(x)[n] = \frac{1}{L} \sum_{k=0}^{L-1} s_{x, \text{L}}[k] \cdot e^{i \frac{2\pi k n}{L}}, \quad n = 0, 1, \dots, L-1.
\label{eq:idft}
\end{equation}

The high-frequency component is obtained by subtracting the low-frequency reconstruction from the original input: \( \text{H}(x) = x - \text{L}(x) \). This decomposition separates fine-grained temporal cues that are often encoded in high-frequency signals.

To reconstruct the final representation, we combine the low-frequency component with a scaled high-frequency counterpart:

\begin{equation}
x_{\text{FFT}} = \text{L}(x) + \beta^2 \cdot \left(x - \text{L}(x)\right),
\end{equation}
where \(\beta\) is a learnable parameter that adaptively controls the contribution of high-frequency information and the square operation ensures that the result is non-negative. This weighting mechanism enables the model to emphasize action-relevant cues while suppressing irrelevant or redundant background patterns introduced by pre-trained features. By modulating the frequency composition, the Global Frequency Decoupling module enhances the model’s focus on atomic action features and improves temporal boundary precision in complex and noisy environments.
\subsubsection{Local High-Frequency Enhancement}

While the Discrete Fourier Transform (DFT) enables global modeling of temporal frequency patterns, it is limited in capturing fine-grained local variations that are crucial for identifying precise action boundaries. To address this, we introduce the Local High-Frequency Enhancement (LHFE) module, which is designed to detect rapid local changes in feature representations and enhance semantic details in boundary regions.

LHFE operates by comparing each frame-level feature with its surrounding temporal context to highlight local deviations. The output of LHFE is computed as:
\begin{equation}
x_{\text{LHFE}} = \sigma \left( \sum_{i=0}^{k-1} w_i \cdot (x_t - \frac{1}{p} \sum_{i=t}^{t+p-1} x_i)_{t+i} \right) + x.
\end{equation}
where \(p\) is the window size used for local averaging, which captures short-term temporal context via mean pooling. The kernel size \(k\) defines the temporal receptive field, and \(w_i\) are learnable weights that control the importance of each local offset. The activation function \(\sigma(\cdot)\) introduces non-linearity.

LHFE complements the global frequency modeling of the GFD module by refining local semantic details. As shown in Fig.~\ref{fig:fgaad}, the output of LHFE is fused with the global frequency-guided features, and the combined representation is passed to downstream modules for further processing.

\subsection{Long-Short-Term Category-Aware Relation Network}

Current TAD methods face challenges in balancing long-term dependencies and fine-grained boundary detection. To address this, we propose the Long-Short-Term Category-Aware Relation (TCAR) module, which refines the decoupled features by integrating global temporal context and local semantic details. As illustrated in Figure~\ref{fig:tacr}, TCAR combines a bidirectional state-space mechanism with multi-scale convolutional encoding, enhancing both action recognition and boundary localization.

Given the decoupled feature representation \( x \in \mathbb{R}^{B \times L \times D} \) produced by the FGAAD module, TCAR processes it as follows:

\begin{figure*}[t] 
    \centering
    \includegraphics[clip, trim=0.1cm 0.5cm 0.1cm 0.5cm, width=0.8\columnwidth]{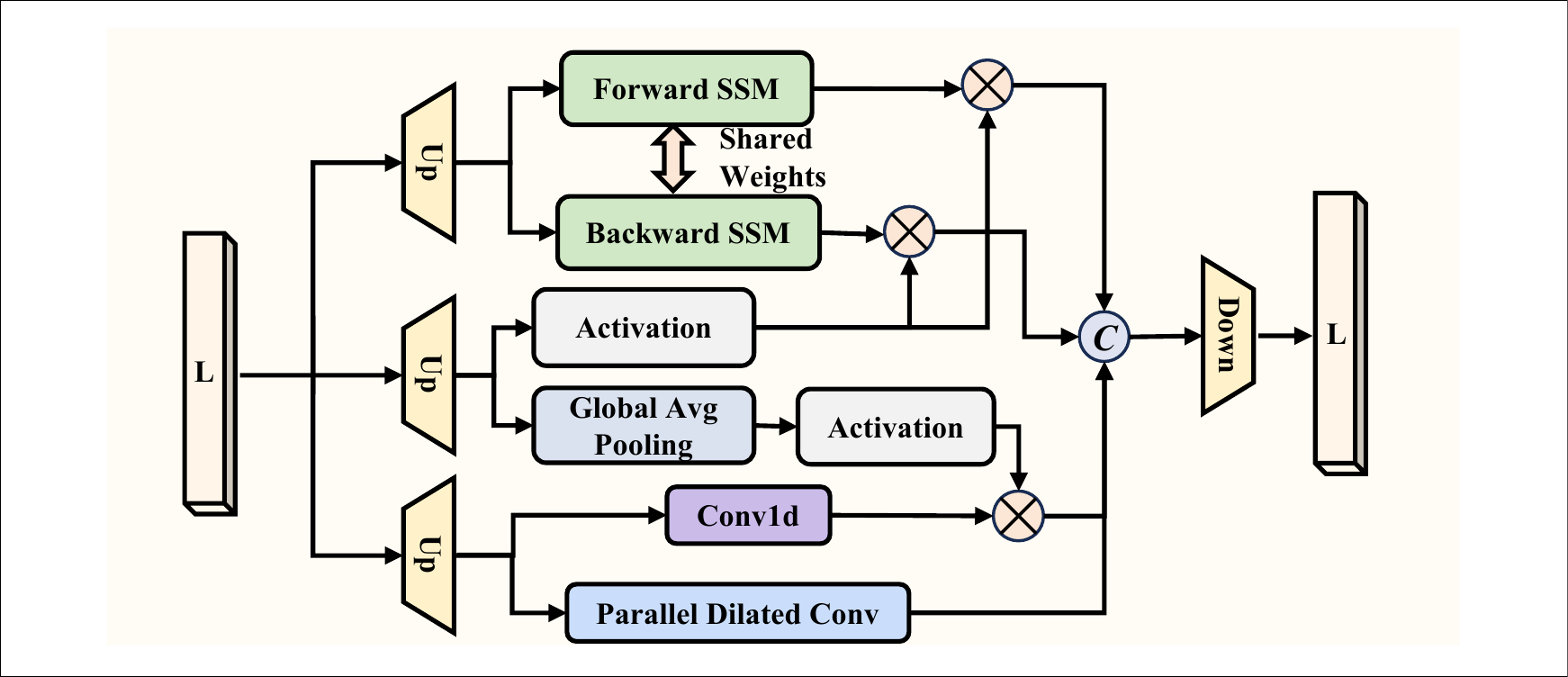}
    \caption{Overview architecture of TACR Block. The "up" and "down" refer to dimensionality increase and reduction, respectively, both performed via linear transformation.}
    \label{fig:tacr}
\end{figure*}

\subsubsection{Bidirectional State-Space Branch}  
To capture long-range and bidirectional dependencies, we adopt a bidirectional State Space Model (SSM)~\cite{ssm,mamba,actionmamba}. At each time step \( t \), the forward and backward hidden states are denoted by \( h_t^{+} \) and \( h_t^{-} \), respectively. The intermediate representation is computed as:
\begin{equation}
z_t^{\pm} = W_h h_t^{\pm} + W_x x_t,
\end{equation}
\begin{equation}
f_t^{\pm} = \text{SiLU}(W_z z_t^{\pm} + W_r x_t),
\end{equation}
where \( W_h, W_x, W_z, W_r \in \mathbb{R}^{D^* \times D^*} \) are learnable projection matrices, and \( D^* \) denotes the latent space dimension. The final global feature is obtained by concatenating the forward and backward outputs:
\begin{equation}
f_{\text{global}} = \text{Concat}(f_t^{+}, f_t^{-}).
\end{equation}

This branch enables the model to incorporate complete temporal context from both directions, essential for understanding long-range action structures.

\subsubsection{Scale-Channel Fusion Branch}  
To complement the global temporal modeling capability of the state-space mechanism, we propose the Scale-Channel Fusion Branch.

We apply a set of parallel 1D dilated convolutions with dilation rates \( r \in \{1, 2, 4\} \) to capture local patterns at different temporal scales:

\begin{equation}
f_{\text{temporal}} = \sum_{r \in \{1, 2, 4\}} \text{Conv}_r(x),
\end{equation}
where \( \text{Conv}_r(\cdot) \) denotes a 1D convolution with dilation rate \( r \).

To complement temporal features with global semantics, we apply global average pooling followed by a non-linear transformation and \(1\text{D}\) convolution:

\begin{equation}
f_{\text{channel}} = \text{Conv}_{\text{pool}} \left( \text{GeLU} \left( \text{Pool}(x) \right) \right),
\end{equation}
where \( \text{Pool}(\cdot) \) denotes global average pooling across the temporal dimension.

The global and local features are fused and passed through a Feed-Forward Network (FFN) to produce the final TCAR output:

\begin{equation}
f_{\text{TCAR}} = \text{FFN} \left( \mathcal{C}(f_{\text{global}}, f_{\text{temporal}}, f_{\text{channel}}) \right),
\end{equation}
where \( \mathcal{C}(\cdot) \) denotes channel-wise concatenation of global context, multi-scale temporal features, and channel-aware semantic cues.

The TCAR module is repeated for \( n \) layers at the original temporal resolution. A downsampling operation with a scale factor of 2 is then applied, and the process is repeated \( m \) times to build a multi-scale temporal feature pyramid. This hierarchical design enriches temporal representation across multiple durations, thereby improving the accuracy of action detection and boundary localization.

\subsection{Loss Function}
The model first processes multi-scale feature pyramids output by the previous module, which capture temporal context at different scales. For each scale, the model outputs \( (p(a_t), d_s^t, d_e^t) \) for each time step \( t \), representing the action probability \( p(a_t) \) and the distances to the start and end action boundaries \( (d_s^t, d_e^t) \). The total loss consists of two components at each scale: \( L_{\text{cls}} \), the focal loss\cite{focal} for classification, and \( L_{\text{reg}} \), the DIoU loss\cite{diou} for boundary regression. The overall loss for video \( X \) is computed as:
\begin{equation}
L = \frac{1}{T_+} \sum_{t=1}^{T} \left[ \alpha (1 - p(a_t))^\gamma \log(p(a_t)) + \lambda_{\text{reg}} c_t \left( 1 - \text{DIoU}(d_s^t, d_e^t, \hat{d}_s^t, \hat{d}_e^t) \right) \right],
\end{equation}
 where \( T_+ \) is the number of positive samples, \( c_t \) is an indicator function for positives, \( \lambda_{\text{reg}} \) balances losses, and \( \gamma \) focuses on hard-to-classify samples by reducing the weight of well-classified ones.
\section{Experiments}
\subsection{Experimental settings}
\subsubsection{Datasets}
We evaluate our method on three widely-used temporal action detection benchmarks: THUMOS14\cite{thumos}, HACS\cite{hacs}, and ActivityNet-1.3\cite{activitynet}. ActivityNet-1.3 and HACS are large-scale datasets covering 200 action categories. ActivityNet-1.3 contains 10,024 videos for training and 4,926 for testing, while HACS includes 37,613 training videos and 5,981 testing videos. Both datasets consist of a diverse range of everyday human activities, making them suitable for evaluating generalizable action detection performance. THUMOS14 focuses on 20 sport-related actions and provides 200 untrimmed videos for training and 213 for testing, with a total of 3,007 and 3,358 action instances in the respective sets. Due to its high temporal density and frequent action transitions, THUMOS14 serves as a challenging benchmark for fine-grained temporal localization. 

\subsubsection{Evaluation}
We follow standard evaluation protocols and report mean Average Precision (mAP) at multiple intersection-over-union (IoU) thresholds. For THUMOS14, we compute mAP at IoU thresholds ranging from 0.3 to 0.7 with a step size of 0.1, i.e., [0.3:0.7:0.1].For both ActivityNet-1.3 and HACS, we report mAP at IoU thresholds of 0.5, 0.75, and 0.95. In addition, the average mAP is computed over the range [0.5:0.95:0.05], as commonly adopted in the literature.

\subsection{Implementation Details}
FDDet is trained end-to-end using the AdamW optimizer with a learning rate of \(10^{-4}\) for THUMOS14, and \(10^{-3}\) for ActivityNet-1.3 and HACS. In the FGAAD module, the low-pass filter cutoff frequency \(C\) is set to 7 and the window size in LHFE is set to 3. Following prior work, we stack \(n = 2\) TCAR layers before applying downsampling with a factor of 2, repeated \(m = 5\) times to construct a multi-scale temporal hierarchy. The hyperparameters of the loss function are chosen according to the guidelines of ActionFormer\cite{actionformer}.

All experiments are conducted on an NVIDIA RTX 4090 GPU with mixed-precision training; for HACS, a dual RTX 4090 setup is used. We adopt the state-of-the-art InternVideo2-6B as the default backbone across all datasets. Additionally, for THUMOS14, we report results using I3D features; for ActivityNet-1.3, we also include R(2+1)D features; and for HACS, we provide further evaluation using VideoMAEv2-giant features. The best results are marked in \textbf{bold} and the second-best results are \underline{underlined}.

\begin{table*}[!h]
    \centering
    \caption{Comparison with the SOTA methods on THUMOS14 and ActivityNet-1.3 datasets. TSN\cite{tsn}, I3D\cite{i3d}, Swin Transformer (Swin)\cite{swin}, TSP (R(2+1)D)\cite{tsp} and IV2-6B(Internvideo2-6B)\cite{internvideo2} features are used. Methods marked with * are our re-evaluations due to the absence of official reports, while † indicates official results from InternVideo2.}
    \label{table:comparison_combined}
    \setlength{\tabcolsep}{2pt} 
    \renewcommand{\arraystretch}{1.2} 
    \scriptsize
    \begin{tabular}{l||l|cccccc||l|cccc}
        \toprule
        \multirow{2}{*}{Method} & \multicolumn{7}{c||}{THUMOS14} & \multicolumn{5}{c}{ActivityNet-1.3} \\
        \cmidrule(lr){2-8} \cmidrule(lr){9-13}
        & Feature & 0.3 & 0.4 & 0.5 & 0.6 & 0.7 & Avg. & Feature & 0.5 & 0.75 & 0.95 & Avg. \\
        \midrule
        TCANet \cite{tcanet} & TSN     & 60.6 & 53.2 & 44.6 & 36.8 & 26.7 & 44.3 & TSN      & 52.3 & 36.7 & 6.9  & 35.5 \\
        RTD-Net \cite{rtd-net} & I3D    & 68.3 & 62.3 & 51.9 & 38.8 & 23.7 & 49.0 & I3D      & 47.2 & 30.7 & 8.6  & 30.8 \\
        VSGN \cite{vsgn}    & TSN      & 66.7 & 60.4 & 52.4 & 41.0 & 30.4 & 50.2 & I3D      & 52.3 & 35.2 & 8.3  & 34.7 \\
        AFSD \cite{afsd}    & I3D      & 67.3 & 62.4 & 55.5 & 43.7 & 31.1 & 52.0 & I3D      & 52.4 & 35.2 & 6.5  & 34.3 \\
        TadTR \cite{tadtr}  & I3D      & 74.8 & 69.1 & 60.1 & 46.6 & 32.8 & 56.7 & TSN      & 51.3 & 35.0 & 9.5  & 34.6 \\
        TALLFormer \cite{tallformer} & Swin & 76.0 & -    & 63.2 & -    & 34.5 & 59.2 & Swin     & 54.1 & 36.2 & 7.9  & 35.6 \\
        ActionFormer \cite{actionformer} & I3D & 82.1 & 77.8 & 71.0 & 59.4 & 43.9 & 66.8 & R(2+1)D & 54.7 & 37.8 & 8.4  & 36.6 \\
        ActionFormer† \cite{actionformer} & IV2-6B & - & - & - & - & - & 72.0 & IV2-6B & - & - & -  & 41.2 \\
        TemporalMaxer* \cite{temporalmaxer} & IV2-6B & \underline{87.9} & 81.3 & 77.3 & 68.2 & 48.3 & 72.6  & IV2-6B & 61.5 & 40.5 & 2.0  & 38.9 \\
        ActionMamba \cite{actionmamba}  & IV2-6B & 86.9 & \textbf{83.1} & 76.9 & 65.9 & 50.8 & 72.7 & IV2-6B      & \underline{62.4} & \underline{43.5} & 10.2  & \underline{42.0} \\
        TriDet \cite{tridet} & I3D     & 83.6 & 80.1 & 72.9 & 62.4 & 47.4 & 69.3 & R(2+1)D & 54.7 & 38.0 & 8.4  & 36.8 \\
        TriDet* \cite{tridet} & IV2-6B     & 87.8 & 81.5 & \underline{78.0} & \underline{68.8} & \underline{52.9} & \underline{73.8} & IV2-6B & 61.3 & 42.7 & 10.3  & 41.4 \\
        DyFADet \cite{dyfadet} & I3D & 84.0 & 80.1 & 72.7 & 61.1 & 47.9 & 69.2 & R(2+1)D & 58.1 & 39.6 & 8.4  & 38.5 \\
        DyFADet* \cite{dyfadet} & IV2-6B & 87.1 & 81.1 & 76.3 & 68.6 & 51.9 & 73.0 & IV2-6B & 61.2 & 42.3 & \underline{10.5}  & 41.3 \\
        \midrule
        \rowcolor{gray!20} Ours(FDDet) & I3D & 83.5 & 79.9 & 72.8 & 62.7 & 48.2 & 69.4 & R(2+1)D & 57.8 & 39.3 & 8.7  & 38.2 \\
        \rowcolor{gray!20} Ours(FDDet) & IV2-6B & \textbf{88.1} & \underline{82.9} & \textbf{78.4} & \textbf{69.7} & \textbf{52.9} & \textbf{74.4} & IV2-6B & \textbf{62.6} & \textbf{43.7} & \textbf{10.6}  & \textbf{42.4} \\
        \bottomrule
    \end{tabular}
\end{table*}

\subsection{Main Results}
\subsubsection{THUMOS14 and ActivityNet-1.3}
The experimental results comparing the performance of our method with other TAD models are shown in Table \ref{table:comparison_combined}. On the THUMOS14 dataset, our method, using I3D features, surpasses the previous state-of-the-art (SOTA) performance. When utilizing the advanced backbone InternVideo2-6B, our method, FDDet, significantly outperforms other TAD methods, surpassing ActionMamba, which uses the same backbone, by approximately 1.7\%, achieving a mAP of \textbf{74.4\%}. We also tested our method using InternVideo2-6B on the previous methods, and our approach still shows a considerable lead. On the ActivityNet-1.3 dataset, we achieved similar results as on THUMOS14, with a mAP of \textbf{42.2\%}. The high performance of the proposed FDDet demonstrates the effectiveness of the frequency-domain decoupling method for TAD, and when using a more advanced backbone, FDDet benefits more compared to other TAD detectors, indicating its superior ability to handle rich semantic features.

\begin{table}[!t]
    \centering
    \caption{Comparison with SOTA Methods on HACS. SlowFast\cite{slowfast}, Swin Transformer (Swin)\cite{swin}, VM2-g(VideoMAEv2-g)\cite{videomaev2} and IV2-6B(Internvideo2-6B)\cite{internvideo2} features are used.}
    \label{table:hacs}
    \setlength{\tabcolsep}{2pt} 
    \renewcommand{\arraystretch}{1.2} 
    \begin{tabular}{l|l|cccc}
    \toprule
    Method & Feature & 0.5 & 0.75 & 0.95 & Avg.  \\
    \midrule
    ActionFormer (ECCV2022) \cite{actionformer} & SlowFast & 54.9 & 36.9 & 9.5 & 36.5 \\
    TALLFormer (ECCV2022) \cite{tallformer} & Swin & 55.0 & 36.1 & 11.8 & 36.5 \\
    TCANet (CVPR2021)\cite{tcanet} & SlowFast & 56.7 & 39.3 & 11.7 & 38.6 \\
    TriDet (CVPR2023) \cite{tridet} & VM2-g & 62.4 & 44.1 & 11.3 & 43.1 \\
    DyFADet (ECCV2024) \cite{dyfadet} & VM2-g & \underline{64.0} & 44.8 & \underline{14.1} & 44.3 \\
    ActionMamba \cite{actionmamba} & IV2-6B & \textbf{64.0} & \underline{45.7} & 13.3 & 44.6 \\
    \midrule
    \rowcolor{gray!20} Ours(FDDet) & VM2-g & 63.5 & 45.1 & \textbf{14.5} & \underline{44.6} \\
    \rowcolor{gray!20} Ours(FDDet) & IV2-6B & 63.9 & \textbf{45.9} & 13.6 & \textbf{44.8} \\
    \bottomrule
    \end{tabular}
\end{table}

\subsubsection{HACS} As shown in Table \ref{table:hacs}, on this dataset, our method continues to maintain its advantage, achieving strong performance across different backbone networks. Notably, our method achieves the highest performance with InternVideo2-6B, setting a new HACS SOTA with a \textbf{44.8\%} mAP, surpassing the previous SOTA, DyFADet (VM2-g). Both VM2-g and IV2-6B are backbones pre-trained on vast amounts of data, each possessing rich semantic representations. The superior performance of FDDet on these two backbones highlights the robustness of our method, demonstrating its ability to effectively adapt to and leverage the rich semantics provided by different pre-trained models.

\subsection{Ablation Study}
We mainly conduct experiments on the THUMOS14 dataset using InternVideo2-6B features to explore more characteristics of FDDet, with a single RTX 4090 GPU.
\subsubsection{Main components analysis}
To evaluate the effectiveness of our proposed components, FGAAD and TCAR, we conduct ablation experiments on the THUMOS14 dataset using InternVideo2-6B features. ActionFormer is selected as the baseline due to its anchor-free and single-stage framework. As presented in Table~\ref{tab:component_analysis}, the integration of FGAAD and TCAR yields consistent improvements across all tIoU thresholds (0.3, 0.5, and 0.7). Specifically, incorporating only FGAAD into the baseline increases the average mAP from {72.1\%} to {73.6\%}. This improvement may be explained by FGAAD’s ability to refine action-relevant features, likely by filtering out irrelevant background noise embedded in pre-trained feature representations. Similarly, the inclusion of only TCAR achieves an average mAP of {73.3\%}, which suggests that TCAR effectively balances global and local temporal dependencies, thereby improving the model’s capacity to capture complex temporal relationships in video sequences. When both components are combined, the full version of our method, FDDet, achieves an average mAP of \textbf{74.4\%}. These results imply that FGAAD and TCAR complement each other, collectively contributing to robust and accurate action detection performance.
\begin{table*}[ht]
    \centering
    \label{tab:combined_analysis}
    
    \begin{minipage}{0.44\textwidth} 
        \centering
        \caption{Effectiveness of components on THUMOS14.}
        \label{tab:component_analysis}
        \setlength{\tabcolsep}{1.5pt} 
        \renewcommand{\arraystretch}{1.1} 
        \begin{tabular}{c|c c|c c c|c}
            \toprule
            Method & FGAAD & TCAR & 0.3 & 0.5 & 0.7 & Avg. \\
            \midrule
            Baseline &            &             & 86.7 & 76.1 & 50.8 & 72.1 \\
            FDDet* & $\checkmark$ &             & \underline{87.6} & \underline{77.6} & \underline{52.1} & \underline{73.6} \\
            FDDet* &            & $\checkmark$ & 87.4 & 77.3 & 51.8 & 73.3 \\
            \midrule
            \rowcolor{gray!20}FDDet & $\checkmark$ & $\checkmark$ & \textbf{88.1} & \textbf{78.4} & \textbf{52.9} & \textbf{74.4} \\
            \bottomrule
        \end{tabular}
    \end{minipage}
    \hfill
    \begin{minipage}{0.44\textwidth} 
        \centering
        \caption{Training speed comparison.}
        \label{tab:speed_analysis}
        \setlength{\tabcolsep}{1.5pt} 
        \renewcommand{\arraystretch}{1.1} 
        \begin{tabular}{l|c|c}
            \toprule
            Method & mAP & Epoch Time \\
            & Avg. & (s) \\
            \midrule
            ActionFormer & 72.0 & 15.6 \\
            TriDet & \underline{73.8} & \textbf{10.3} \\
            DyFADet & 73.0 & 18.3 \\
            \midrule
            \rowcolor{gray!20}FDDet & \textbf{74.4} & \underline{11.1} \\
            \bottomrule
        \end{tabular}
    \end{minipage}
\end{table*}

\subsubsection{Training Efficiency Analysis}
\begin{figure}[t]  
    \centering        
    \includegraphics[width=0.8\textwidth]{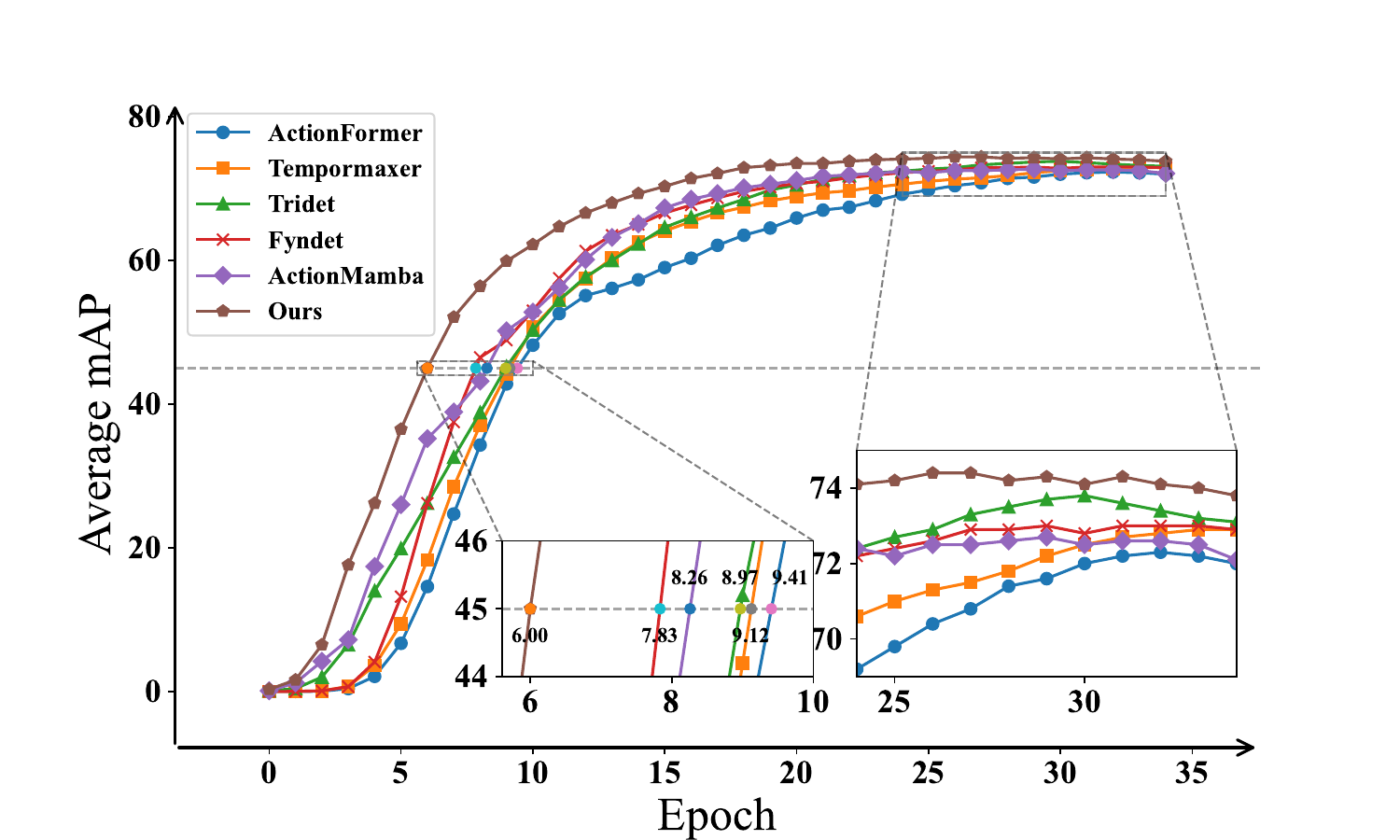} 
    \caption{This chart illustrates the training efficiency of various SOTA models on the THUMOS14 dataset using InternVideo2-6B features.} 
    \label{fig:zoomed_plot} 
\end{figure}

As shown in Table \ref{tab:speed_analysis} and Figure\ref{fig:zoomed_plot}, we analyze both the time spent per epoch and the progression of mAP throughout the epochs. Our method outperforms the previous SOTA in terms of training time per epoch, demonstrating a clear advantage in training efficiency. Furthermore, our approach achieves strong performance with fewer epochs, highlighting both its efficiency and effectiveness compared to other methods.
In the early stages of training, our method quickly surpasses the others, due to the ability of FGAAD to refine action-relevant features by suppressing background noise. In the later stages, it continues to maintain robust performance, benefiting from TCAR’s capability to capture both global and local temporal dependencies, ensuring consistent and accurate action localization.

\begin{figure}[ht]
    \centering
    \begin{minipage}{0.48\textwidth} 
        \centering
        \includegraphics[width=\textwidth]{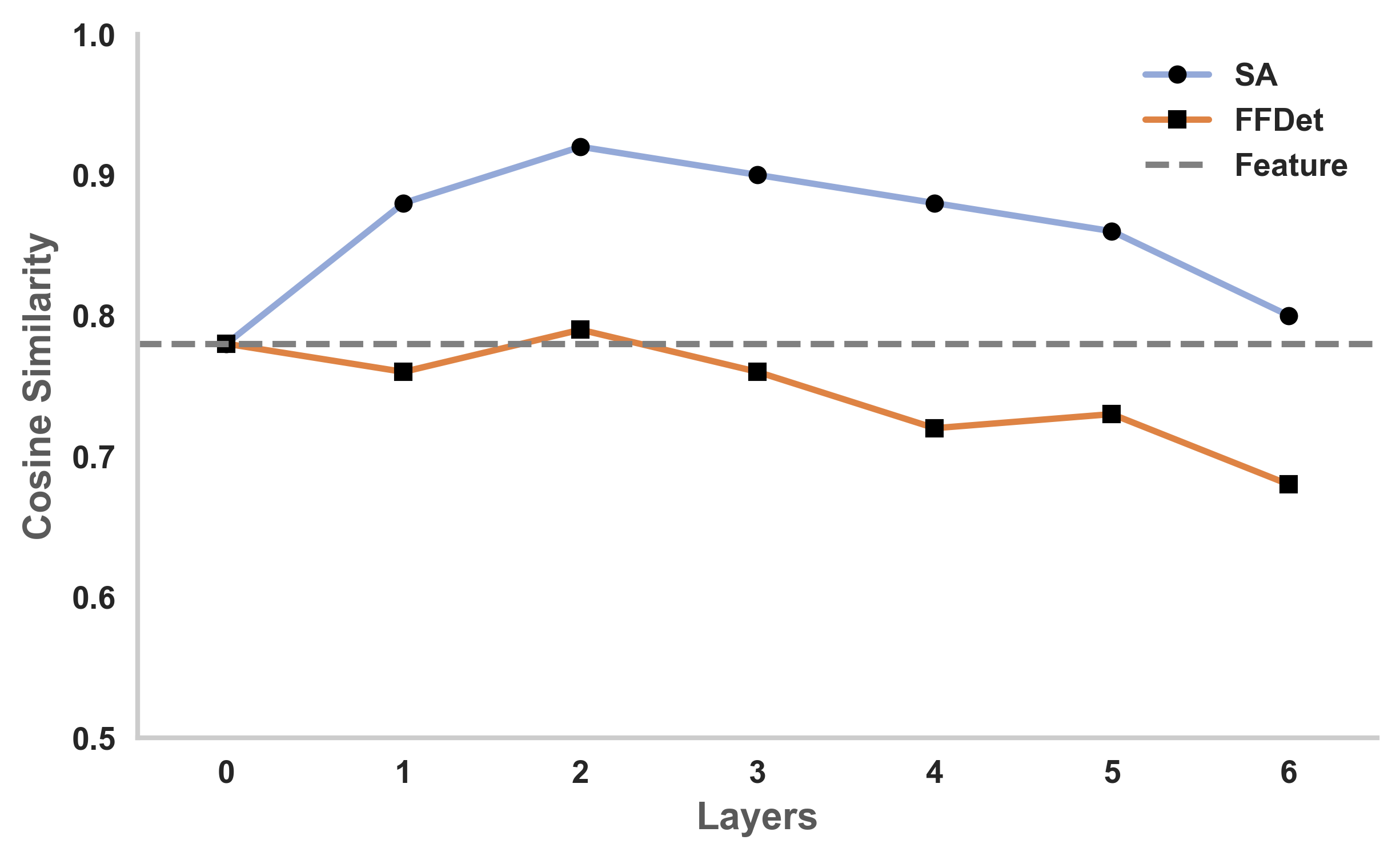}
        \caption{Average cosine similarity between features at each layer.}
        \label{fig:cos}
    \end{minipage} \hfill  
    \begin{minipage}{0.48\textwidth} 
        \centering
        \includegraphics[width=\textwidth]{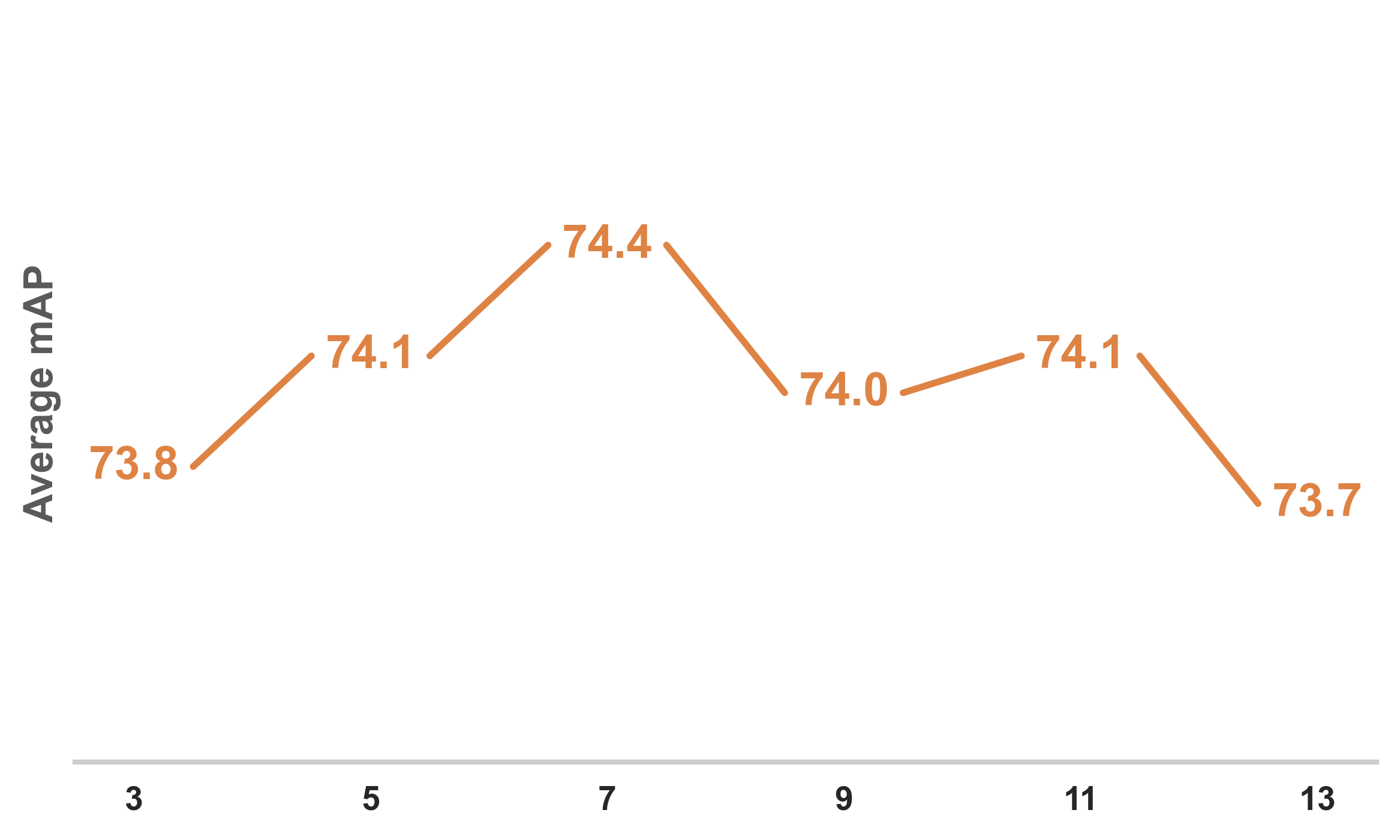}
        \caption{Effectiveness of the Cutoff frequency c.}
        \label{fig:value_c}
    \end{minipage}
    \label{fig:main}
\end{figure}

\subsubsection{The distinguishability of the learned features} As shown in Figure~\ref{fig:cos}, the features obtained by recent TAD methods\cite{actionformer} tend to exhibit high similarities, making it difficult for the model to precisely locate action boundaries. The high similarity in these features is likely due to the fact that features extracted from large pre-trained models often focus on global image context. As a result, during action transitions, the feature variations are minimal, making it harder to differentiate actions. In contrast, after FGAAD feature enhancement, our model effectively filters out irrelevant background semantics, leading to a significant reduction in feature similarity. Furthermore, with the support of the TCAR layer, the features retain their high distinguishability, indicating that our frequency-domain enhancement effectively improves the model's ability to capture the correct responses to action-related cues.

\subsubsection{Ablation on the cutoff frequency in FGAAD} In this section, we examine the impact of different cutoff frequencies on FDDet performance. As shown in Figure~\ref{fig:value_c}, the model achieves its best performance when the cutoff frequency is set to 7, reaching an average mAP of 74.4\%. Beyond this point, the performance remains relatively stable but begins to slightly decline. These results suggest that a cutoff frequency of 7 provides the best trade-off between retaining discriminative action features and removing redundant background information, making it a reliable default setting for the FGAAD module.

\subsection{Error Analysis}

To better understand the behavior of our model, we perform comprehensive error analysis on the THUMOS14 test set using the diagnostic toolkit provided by~\cite{detad}. Based on InternVideo2-6B features, our model is evaluated in terms of false negative distributions, sensitivity to different action characteristics, and fine-grained false positive types. These analyses provide insights beyond standard mAP scores. For evaluation protocol and metric definitions, please refer to~\cite{detad}.

\subsubsection{Characteristic Metrics} 
Following~\cite{detad}, we adopt three action-level characteristics for detailed analysis: coverage, length, and number of instances. Coverage measures the proportion of the video occupied by an action instance, computed as the ratio of the action duration to the full video length. It is categorized into five bins: Extra Small (XS: $(0, 0.02]$), Small (S: $(0.02, 0.04]$), Medium (M: $(0.04, 0.06]$), Large (L: $(0.06, 0.08]$), and Extra Large (XL: $(0.08, 1.0]$). Length refers to the absolute duration of an action instance, measured in seconds. It is divided into five ranges: XS $(0, 3]$, S $(3, 6]$, M $(6, 12]$, L $(12, 18]$, and XL $(>18)$. Number of instances indicates the total count of action instances from the same class within a video. This reflects the density of action occurrences and is grouped as XS $(1)$, S $[2, 40]$, M $[40, 80]$, and L $(>80)$. 

\begin{figure}[ht]
    \centering
    \includegraphics[clip, trim=0.1cm 0.2cm 0.1cm 0.2cm, width=0.9\linewidth]{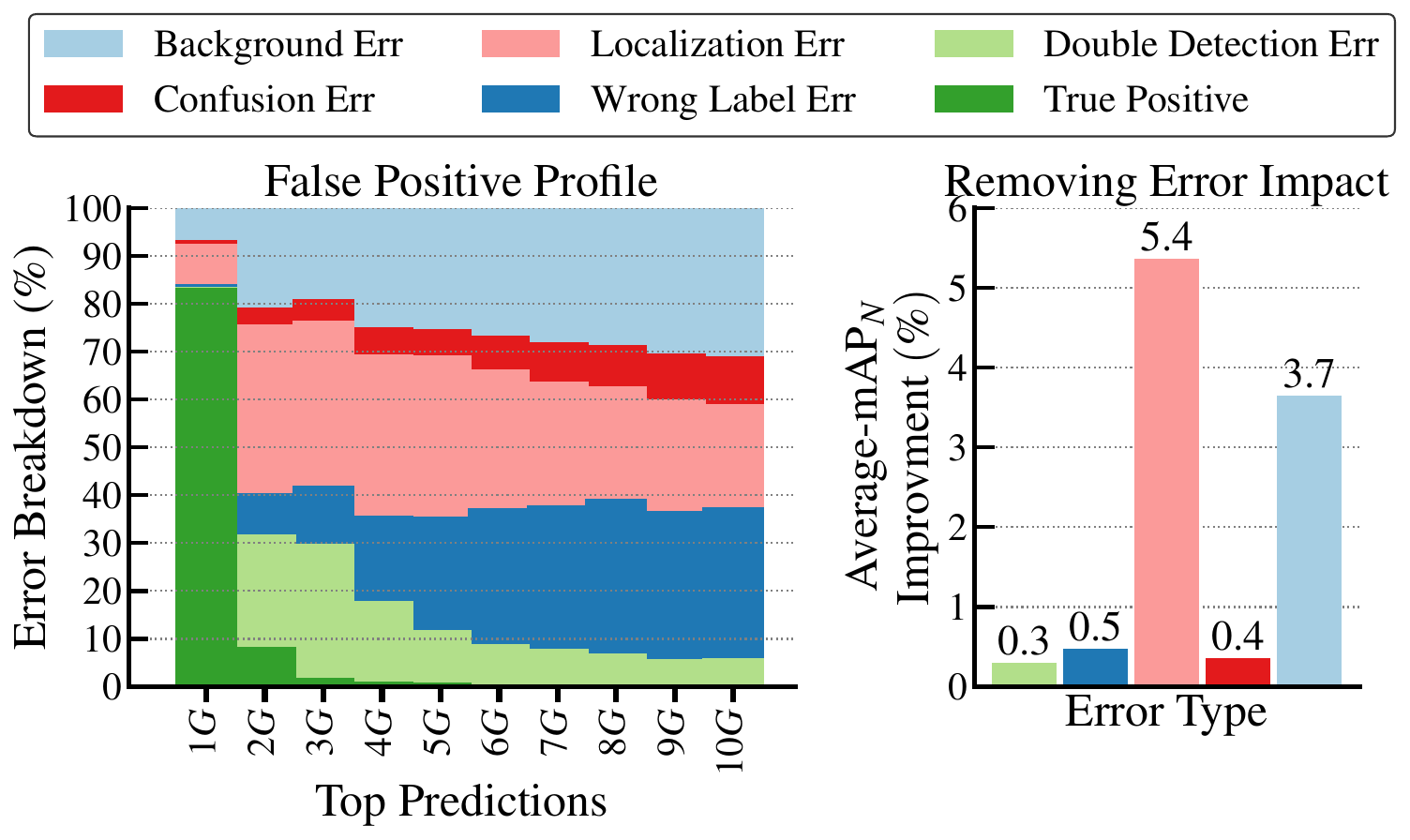}
    \caption{False positive analysis: distribution of error types across prediction ranks (left), and estimated average mAP$_N$ improvement by removing each type (right).}
    \label{fig:analysis_fp}
\end{figure}
\subsubsection{False Positive Analysis}
Fig.~\ref{fig:analysis_fp} presents the distribution of different false positive types under increasing numbers of top-ranked predictions, where $G$ represents the number of ground-truth instances per class and the top $k \times G$ predictions are retained for evaluation. In the leftmost 1$G$ column, true positives make up about 83\% of predictions at $\text{tIoU}=0.5$, demonstrating that our model effectively ranks high-quality predictions with high confidence. As $k$ increases, localization and background errors become more prominent, suggesting that low-quality predictions tend to involve inaccurate boundaries or irrelevant content. The right-hand bar chart summarizes the average mAP gain obtained by removing each error type, revealing that localization and background errors have the largest impact on overall performance.

\begin{figure}[ht]
    \centering
    \includegraphics[clip, trim=0.1cm 0.2cm 0.1cm 0.3cm, width=\linewidth]{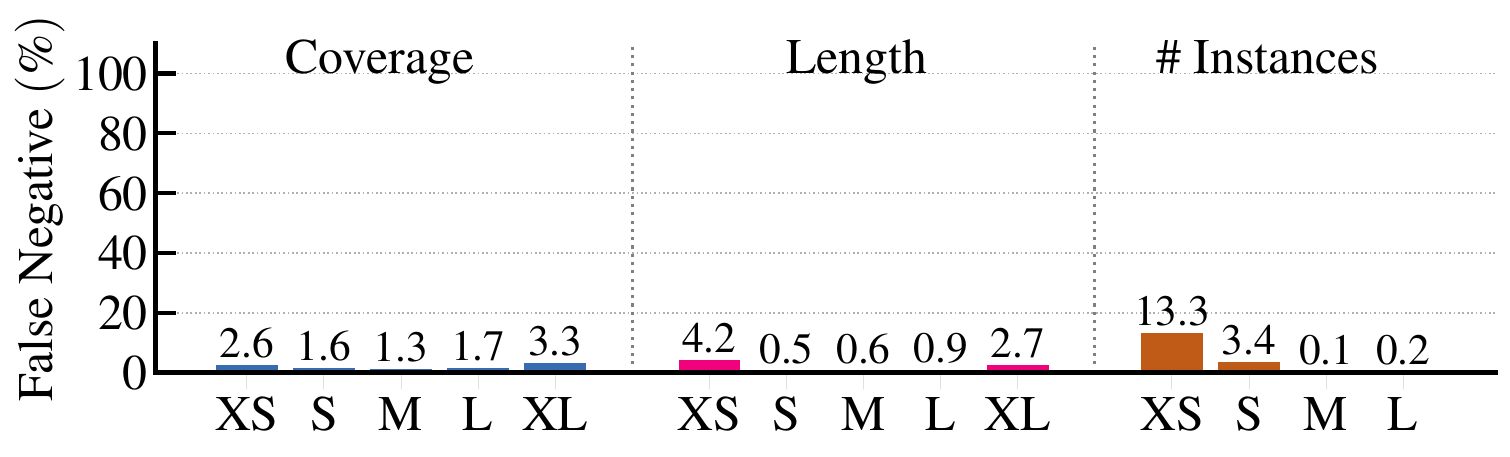}
    \caption{False negative analysis over three action-level characteristics: coverage, length, and number of instances. The figure shows how often the model misses actions under different conditions.}
    \label{fig:analysis_fn}
\end{figure}

\subsubsection{False Negative Analysis} 
Fig.~\ref{fig:analysis_fn} shows the distribution of false negatives across different action characteristics, including coverage, action length, and the number of instances per video. We observe that false negative rates remain relatively consistent across different coverage levels. However, actions with extremely short or long durations (XS and XL in length) are more likely to be missed, possibly due to insufficient temporal context or over-fragmentation. Additionally, videos with fewer action instances also exhibit higher false negative rates, which may be attributed to the lack of dense contextual cues. Notably, our model achieves a lower overall miss rate compared to prior methods such as ActionFormer~\cite{actionformer} and TriDet~\cite{tridet}.

\begin{figure}[ht]
    \centering
    \includegraphics[clip, trim=0.1cm 0.2cm 0.1cm 0.2cm, width=\linewidth]{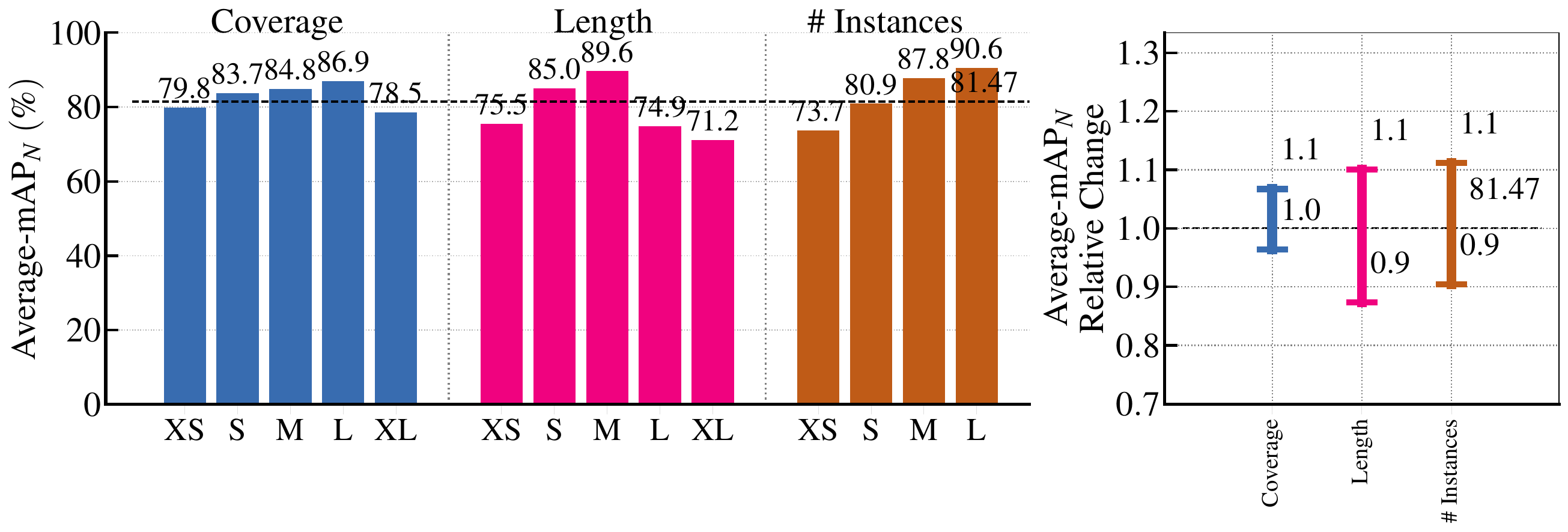} 
    \caption{Sensitivity analysis: average mAP$_N$ grouped by different characteristics (left), and relative change with standard deviation (right).}
    \label{fig:analysis_sensitivity}
\end{figure}

\subsubsection{Sensitivity Profiling} 
Fig.~\ref{fig:analysis_sensitivity} shows the average mAP$_N$ under different action characteristics, including coverage, length, and number of instances per video. Our model performs best on actions with moderate coverage and duration (S/M/L), while detection accuracy drops for extremely short or long actions (XS/XL). However, videos with more action instances tend to achieve higher mAP, possibly due to stronger contextual cues, whereas sparse-action videos remain more challenging. Overall, our model exhibits stable performance across conditions and surpasses prior methods.

\section{Conclusion}
In this paper, we propose a novel method for Temporal Action Detection (TAD) that addresses the limitations of frozen feature extraction by introducing a frequency-aware decoupling network and a temporal category-aware relation network. Our approach refines action-relevant features by filtering out irrelevant background noise and balancing global and local temporal dependencies, which leads to improved boundary precision and action localization. Experiments on three benchmark datasets, THUMOS14, ActivityNet-1.3, and HACS, demonstrate that our method achieves sota performance in diverse scenarios, highlighting its robustness and generalizability. Ablation studies further validate the effectiveness of our key components: the Frequency-Guided Atomic Action Decoupling Network refines discriminative features by suppressing irrelevant semantics, while the Long-Short-Term Category-Aware Relation Network enhances temporal modeling by capturing both global and local dependencies. Additional ablation experiments further support the robustness of our approach, showing consistent improvements in accuracy and convergence speed. These results underscore the efficiency, adaptability, and superior performance of our framework to address various challenges in TAD tasks.

%
%
%

\bibliographystyle{splncs04}

\bibliography{references}

%




\end{document}